\pdfoutput=1

\documentclass[11pt]{article}

\usepackage[]{acl}

\usepackage{times}
\usepackage{latexsym}

\usepackage[T1]{fontenc}

\usepackage[utf8]{inputenc}

\usepackage{microtype}
\usepackage[frozencache,cachedir=.]{minted}
\usepackage{graphicx}

\title{Performance of Data Augmentation Methods for Brazilian Portuguese Text Classification}

\author{Marcellus Amadeus \and Paulo Branco \\
  Alana AI \\
  \texttt{\{marcellus\}@alana.ai} \\}

\begin{document}
\maketitle
\begin{abstract}
Improving machine learning performance while increasing model generalization has been a constantly pursued goal by AI researchers. Data augmentation techniques are often used towards achieving this target, and most of its evaluation is made using English corpora. In this work, we took advantage of different existing data augmentation methods to analyze their performances applied to text classification problems using Brazilian Portuguese corpora. As a result, our analysis shows some putative improvements in using some of these techniques; however, it also suggests further exploitation of language bias and non-English text data scarcity.
\end{abstract}

\section{Introduction}

Text classification is a common and essential task in natural language processing (NLP). Much work has been done in the area. State-of-the-art results achieved high accuracy on several related tasks such as sentiment analysis \citep{thongtan2019, jiang2020} and topic classification \citep{santosh2019, meng2020}. Still, high performance often depends on the size, quality, and perhaps most important, training data availability. Gathering data can quickly become a tedious assignment and it is especially challenging for non-English languages that likely have fewer resources since most current researches use English corpora. In such a scarce scenario, data augmentation techniques are even handier to deal with data. Data augmentation is already widely used in computer vision \citep{simard2012, szegedy2015, krizhevsky2017} and speech \citep{cui2015, ko2015}, where it boosts performance, especially on smaller datasets.

Text data augmentation techniques use various strategies such as applying a set of universal functions to quickly and easily introduce diversity in the dataset \citep{wei2019}, generating new data by translating sentences into another language and back into English \citep{yu2018} (also referred to as "back translation"), using predictive language models for synonym replacement \citep{kobayashi2018}, and others. Thus, implementation cost versus performance gain varies from technique to technique. Still, all of the methods rely on at least one kind of language resource, which may be a WordNet dictionary, a Word Embedding model, datasets with specific formats, or another type of dependency closely tied to a single language.

Most of these text augmentation techniques were originally developed using English corpora. However, some recent works extend their application scenarios using a specific technique on, whether various languages \citep{ciolino2021}, or even Brazilian Portuguese corpora as evaluation language \citep{verissimo2020, venturott2020}. Each work uses distinct processes and datasets and, given some limitations, data augmentation improved their results. 

In this paper, we revisit some different existing text augmentation methods, gathering and reconstructing the necessary resources to reproduce each technique. Then, by applying them to Brazilian Portuguese corpora, we attempt to expand and validate these techniques in a more generic way. Using McNemar's statistical test to compare the classification models, we achieve a set of different results showing that text augmentation methods are particularly useful, although language-specific fine-tunings should be considered to ensure significant positive gains.

\section{Experimental Setup}

Initially, we cluster the text augmentation methods into three main groups: (1) \textbf{Easy Data Augmentation} (EDA): based on \citet{wei2019}, it consists of a collection of four functions (synonym replacement, random insertion, random swap, and random deletion). The first two rely on a map of synonyms; in this case, we use the PPDB Portuguese paraphrase pack (available at \url{http://paraphrase.org}). For the other parameters of the technique, we use the same ones in the original paper. (2) \textbf{Synonym} (Syn): Many text augmentation methods are based on some kind of synonym or allonym replacement. They primarily use language models to effectively replace words to create diversity in the training set synthetically. Here we use the \mintinline{Python}{nlpaug} library\footnote{\url{https://github.com/makcedward/nlpaug}} to produce a pipeline of word replacement (sequential flow) that uses the PPDB Portuguese paraphrase pack, the Fasttext Portuguese Word Embedding model\footnote{\url{https://fasttext.cc/docs/en/crawl-vectors.html}} \citep{grave2018}, and the Portuguese BERT model\footnote{\url{https://huggingface.co/neuralmind/bert-large-portuguese-cased}} \citep{souza2019}. Combined, the three resources provide a smart replacement for similar words. We generate one new sentence per sample in the training set using this method. (3) \textbf{Back translation} (BT): many translation APIs are publicly available and free (up to reasonable usage). So for this method, we generate one sentence per sample in the training set using AWS Amazon Translate service, chosen for convenience (Microsoft and Google also offer that kind of API, where it is also possible to use pre-trained models to perform the "back translation" technique).

\subsection{Benchmark Datasets}

We conduct experiments on three public available Brazilian Portuguese text classification datasets: (1) \textbf{Tweets}: TweetSentBR \citep{brum2017} is a corpus of 10,648 Tweets manually annotated in one of the three sentiment classes: Positive - the user meant a positive reaction or evaluation about the main topic on the post; Negative - the user meant a negative reaction or evaluation about the main topic on the post; Neutral - tweets not belonging to any of the last classes, usually not making a point, out of topic, irrelevant, confusing or containing only objective data. (2) \textbf{B2W}: B2W Open Product Reviews (available at \url{https://github.com/b2wdigital/b2w-reviews01}) consists of a binary classification corpus of 132,373 product reviews. The labels represent the willingness of the customer to recommend the product to someone else; and (3) \textbf{Mercadolibre}: Mercado Libre Data Challenge 2019 (available at \url{https://ml-challenge.mercadolibre.com}) is a corpus of 693,318 purchase histories where the goal is to predict the next item bought by the user. 
    
As demonstrated by \citet{wei2019}, some text augmentation techniques have a more significant impact on smaller datasets; for that reason, we randomly resample the datasets into subsets with different sizes \(N=\{500, 1000, 2000, 5000, 10000\}\). For each subset \(N\), we use a 75\% train split rate. Also, we use different percentages \(p=\{0\%, 5\%, 10\%, 20\%\}\) of the train set in the augmentation process. Finally, we run 15 rounds of the whole experiment for each dataset, totaling 2,700 trained models (3 datasets X 3 augmentation groups X 5 subset sizes X 4 augmentation percentages X 15 rounds).

\subsection{Text Classification Models}

Due to a large number of models, we opted to use non-deep-learning classifiers to perform the benchmark based on their ease of use and usually faster training. Many popular algorithms for text classification are not based on neural networks; one of the most prominent is the Support Vector Machine (SVM) algorithm \citep{kowsari2019}. 

We use the sci-kit-learn library\footnote{\url{https://scikit-learn.org/stable/modules/generated/sklearn.svm.SVC.html\#sklearn.svm.SVC}} implementation of SVM with \mintinline{Python}{C=10}, \mintinline{Python}{kernel=rbf}, and \mintinline{Python}{gamma=scale} for all trained models (other parameters set to default). As the featurizer, we use the Fasttext Portuguese Word Embedding model\footnote{\url{https://fasttext.cc/docs/en/crawl-vectors.html}} \citep{grave2018} to extract the sentence vector for each sample. 

\subsection{Model Evaluation}

We use the F1-score (weighted F1-score for multi-label datasets) as the evaluation metric. The F1-score is the harmonic mean of precision and recall, and it was applied as a filter, leaving only the best 180 models for each experiment round and parameter combination. 

After the models' filtering, we use the F1-score of each baseline model (\(p = 0\%\)) to compute the model gain, \emph{i.e.} the difference between the baseline F1-score and the F1-score of each augmented model (\(p > 0\%\)). Also, to determine if a model's performance is significantly different, we use the continuity-corrected version of McNemar's test. This nonparametric statistical method is used on paired nominal data and has been used to confront NLP models \citep{jiang2021, chen2021}.

\section{Results}

All details regarding baseline F1-scores, F1-score gains and \(p\)-values can be found in Appendices \ref{sec:appendix_baseline}, \ref{sec:appendix_gains} and \ref{sec:appendix_pvals}, respectively.


\subsection{Tweets Dataset}

\begin{figure}[h]
    \centering
    \includegraphics[scale=0.31]{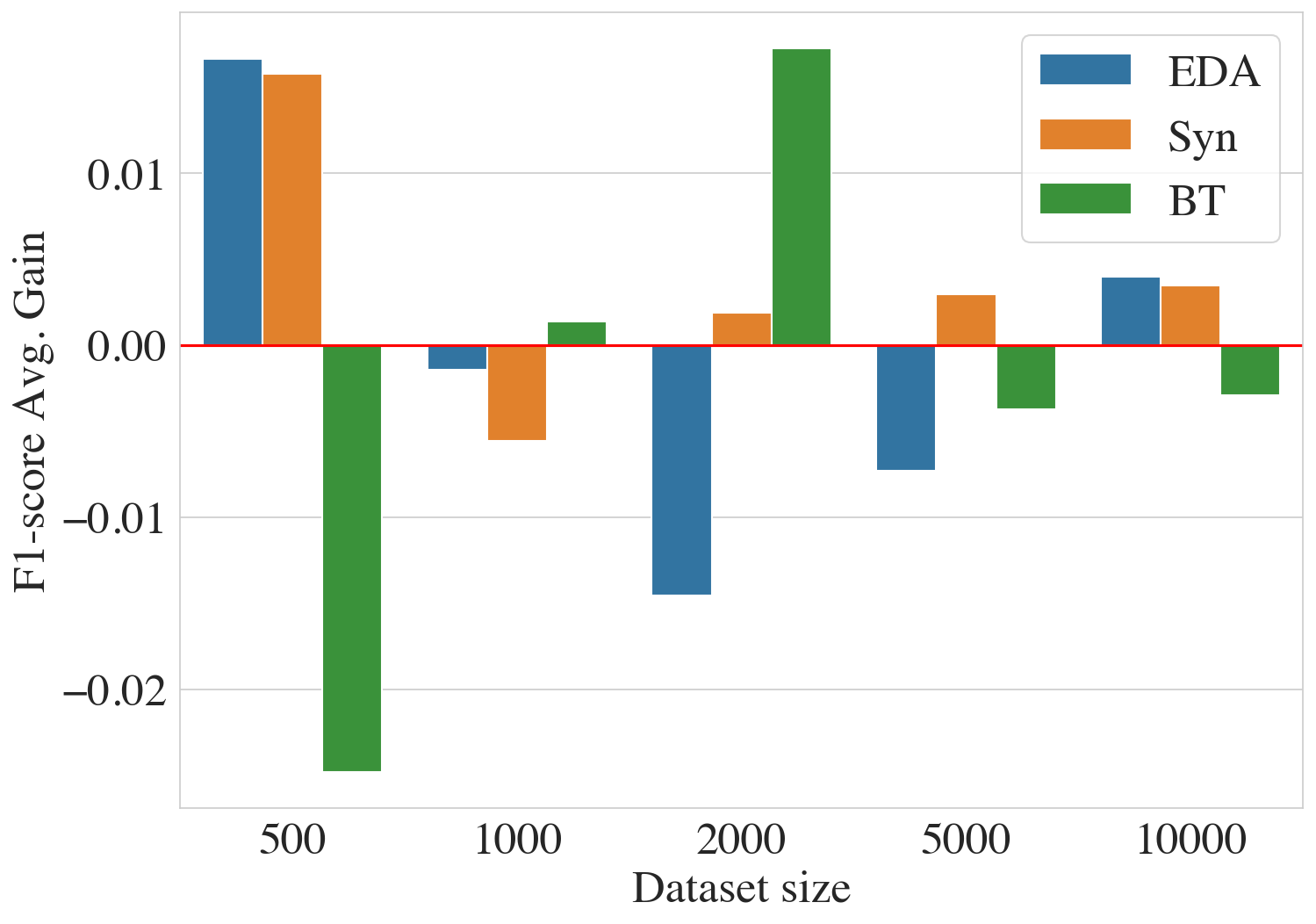}
    \caption{F1-score average gains for each augmentation group on Tweets dataset.}
    \label{fig:tw_avg_gain}
\end{figure}

As a result of the Tweets dataset analysis, the highest F1-score gains were obtained on the smallest dataset subset for EDA and Syn augmentation groups. Figure \ref{fig:tw_avg_gain} shows the F1-score average gains for each augmentation group.

\begin{figure}[h]
    \centering
    \includegraphics[scale=0.37]{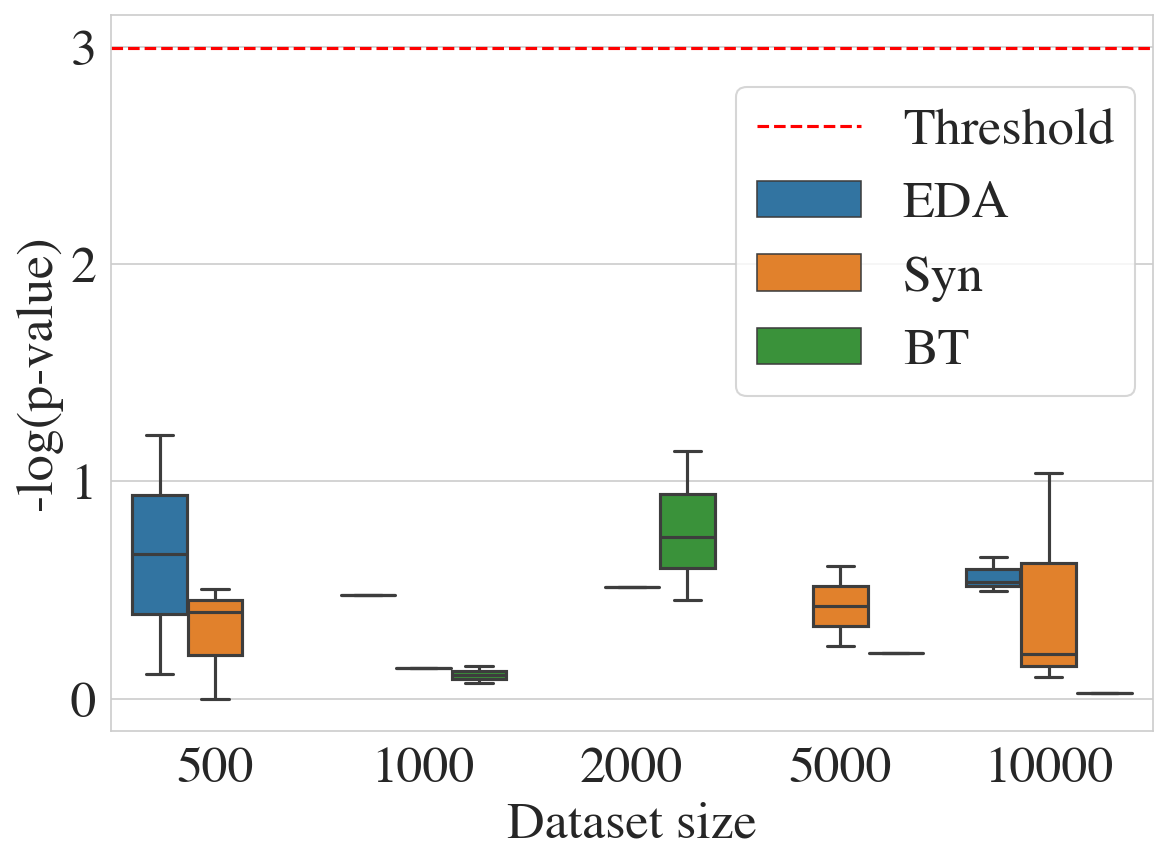}
    \caption{-log(\(p\)-values) for each augmentation group on Tweets dataset. Dashed red line: alpha = 0.05.}
    \label{fig:tw_pval}
\end{figure}

McNemar's test results show no significant difference between baseline models and augmented models (\(p\)-value < 0.05). The test was applied only in positive F1-score gain models and is depicted in Figure \ref{fig:tw_pval}.

\subsection{B2W Dataset}

\begin{figure}[h]
    \centering
    \includegraphics[scale=0.31]{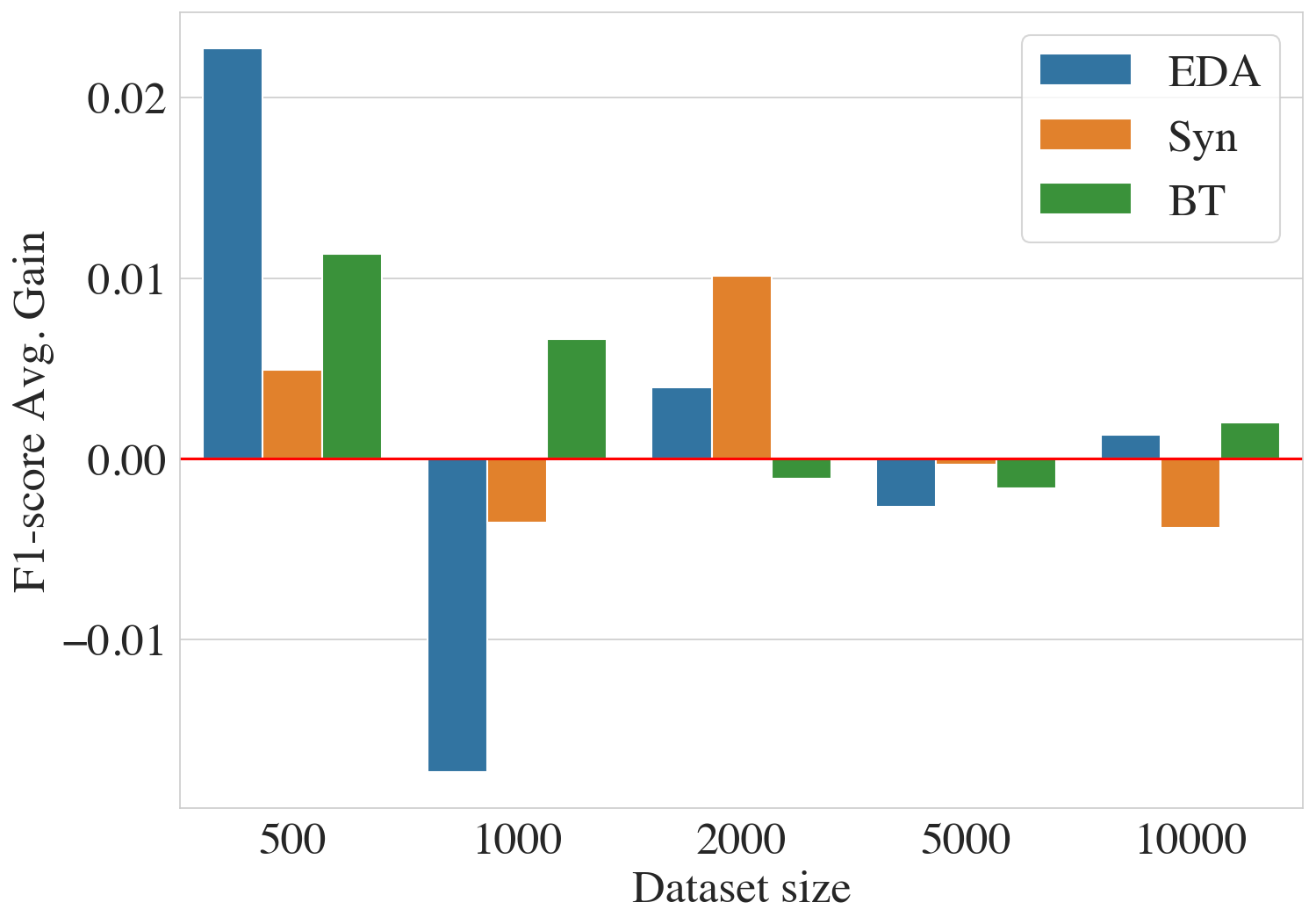}
    \caption{F1-score average gains for each augmentation group on the B2W dataset.}
    \label{fig:bw_avg_gain}
\end{figure}

As with Tweets, the analysis of the B2W dataset resulted in the highest F1-score gain belonging to the EDA augmentation group. Also, the dataset subsets sizes that achieved the best results were 500 and 2000. Likewise, the same pattern was observed in the Tweets dataset. Figure \ref{fig:bw_avg_gain} shows the F1-score average gains for each augmentation group.

\begin{figure}[h]
    \centering
    \includegraphics[scale=0.37]{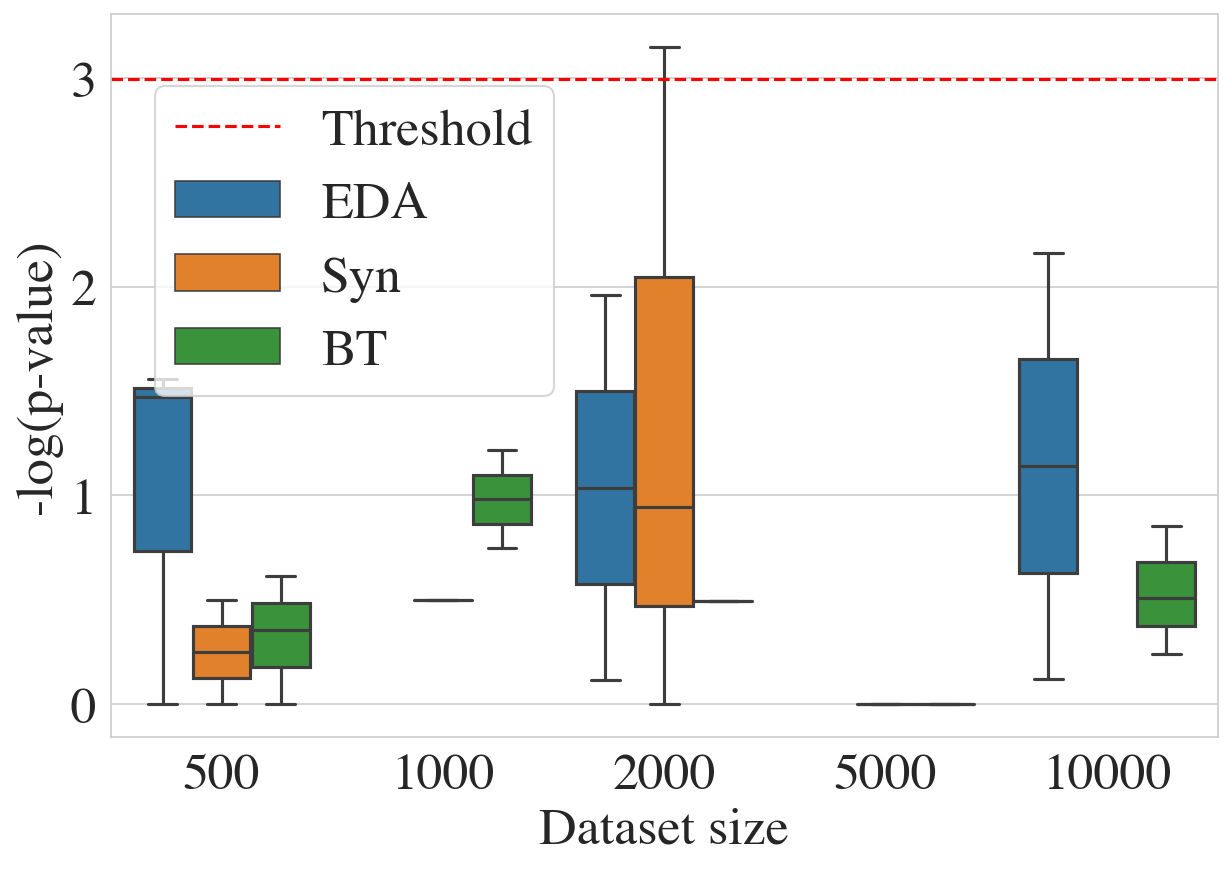}
    \caption{-log(\(p\)-values) for each augmentation group on B2W dataset. Dashed red line: alpha = 0.05.}
    \label{fig:bw_pval}
\end{figure}

The statistical significance analysis was applied following the same scheme as the Tweets dataset. Figure \ref{fig:bw_pval} shows the positive gain models' \(p\)-values.

The B2W dataset analysis resulted in a significant model, as shown in Figure 4. This best model, trained with a subset size 2000 and 0.05 augmentation, reached a \(p\)-value of 0.042957.

\subsection{Mercado Libre Dataset}

\begin{figure}[h]
    \centering
    \includegraphics[scale=0.31]{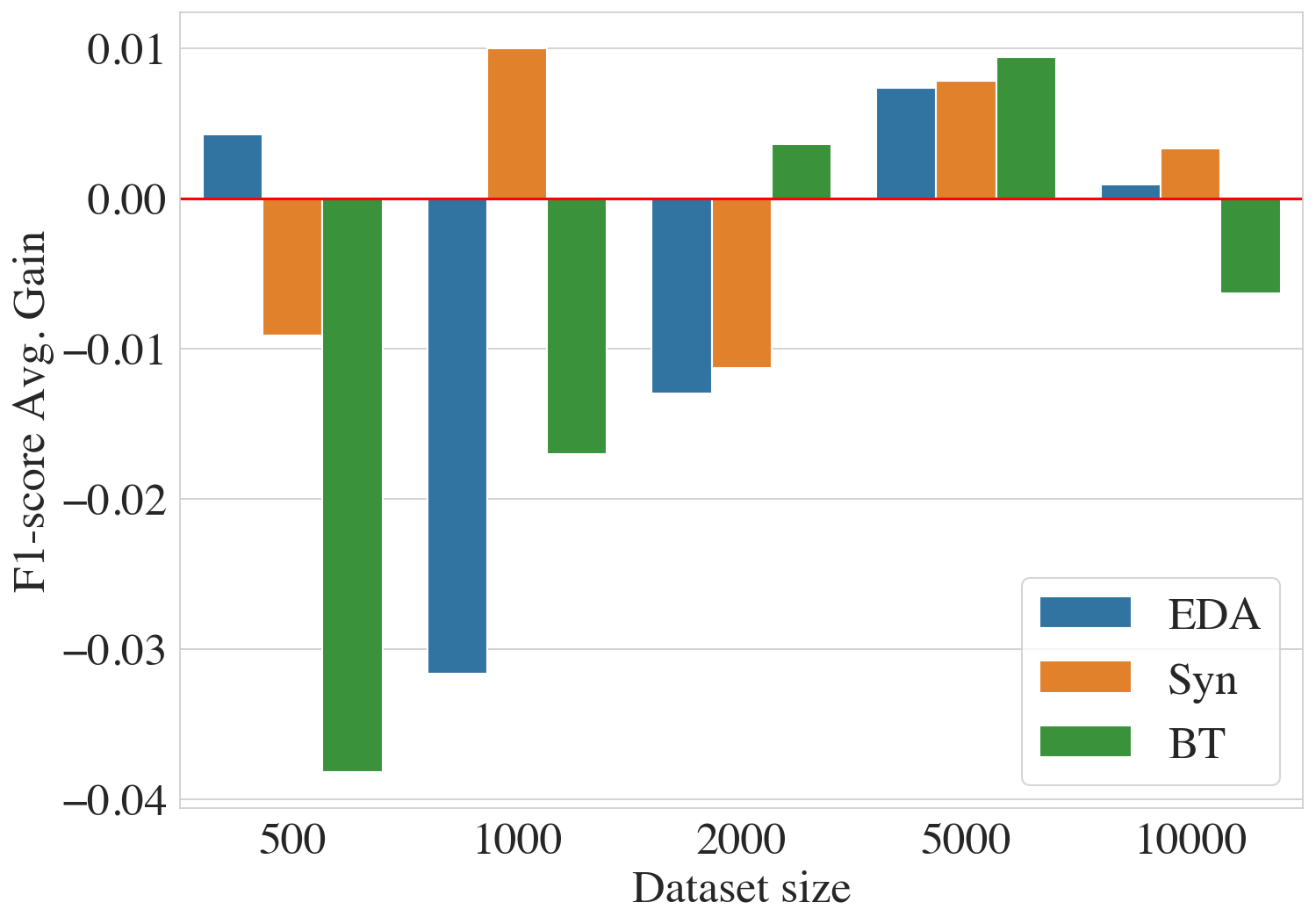}
    \caption{F1-score average gains for each augmentation group on Mercado Libre dataset.}
    \label{fig:ml_avg_gain}
\end{figure}

Finally, in the Mercado Libre dataset, the EDA augmentation group still achieved the best F1-score gain. Figure \ref{fig:ml_avg_gain} shows the F1-score average gain for each group.

\begin{figure}[h]
    \centering
    \includegraphics[scale=0.37]{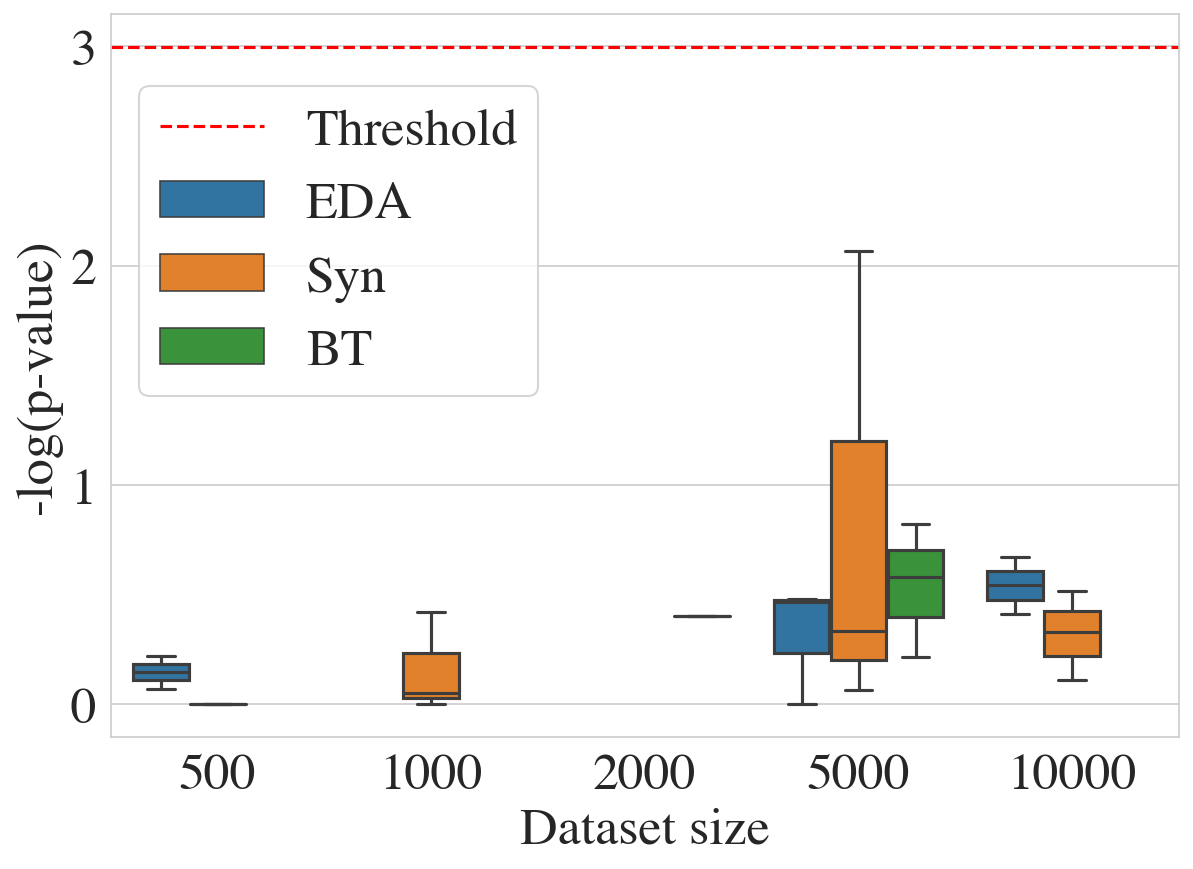}
    \caption{-log(\(p\)-values) for each augmentation group on Mercado Libre. Dashed red line: alpha = 0.05.}
    \label{fig:ml_pval}
\end{figure}

All positive F1-score gains p-values resulting from McNemar's test are depicted in Figure \ref{fig:ml_pval}.

Although no significant text classification model was found (Figure \ref{fig:ml_pval}), the dataset subset size 5000 models obtained a satisfactory performance.

\subsection{Augmentation Group Performance}

\begin{figure}[h]
    \centering
    \includegraphics[scale=0.35]{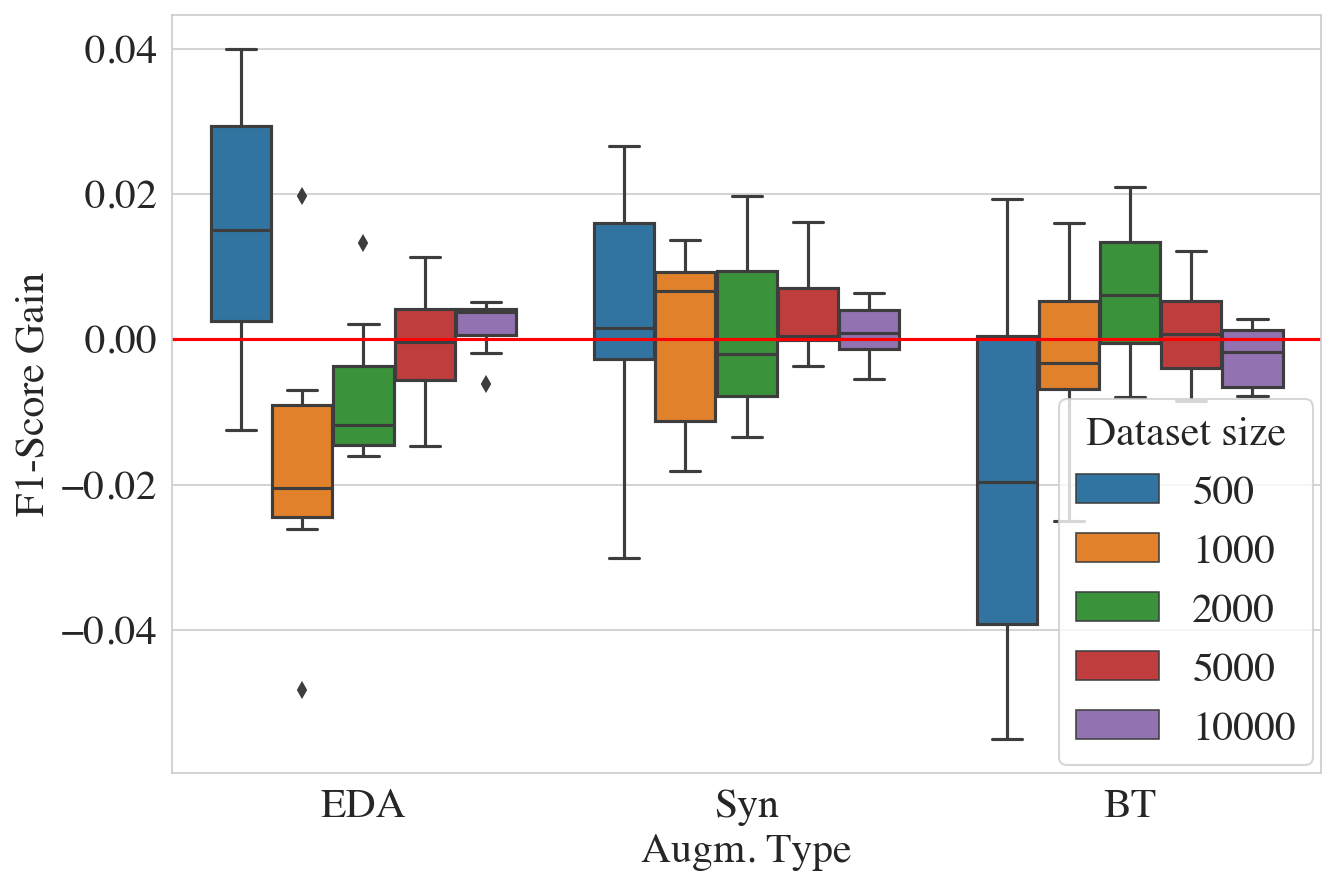}
    \caption{F1-score gains for each augmentation group, combining all datasets.}
    \label{fig:overall_avg_gain}
\end{figure}

Combining all results across the three augmentation groups, Figure \ref{fig:overall_avg_gain} shows the F1-score gains for each dataset subset size.

The Syn augmentation group achieved the best overall performance, as shown in Figure \ref{fig:overall_avg_gain}.

\section{Conclusions}

We grouped and compared three different text augmentation techniques that have been reported as significant additions to solving text classification tasks. Initially developed using English corpora, we applied these techniques using Brazilian Portuguese publicly available language resources.

We trained 2,700 text classification models using three different datasets. For each dataset, the models were generated using combinations of the following attributes: dataset subset size, percentage of augmentation, and the group of augmentation technique.

Comparing augmented and non-augmented best models, our analysis showed a slight upward trend in the F1 score gain, although no expressive statistical significance was found. This result might be caused by the model choice, \emph{i.e.} the SVM is considered a low-sensitive method regarding the train set size; and by the nature of the datasets, \emph{i.e.} the colloquialism inherent in the Tweets texts and a large number of targets in the Mercado Libre classification task, adding noise to the trained models. Also, since all the augmentation techniques were developed using English corpora, some putative language dependency might occur, biasing the final results.

In future work, we are planning to adjust the model choice and expand the analysis on the Brazilian Portuguese language. Besides, in order to increase the available Brazilian Portuguese data volume, we are planning to curate and annotate a new corpus.

\bibliography{anthology,custom}
\bibliographystyle{acl_natbib}

\appendix

\section{Detailed Baseline F1-scores}
\label{sec:appendix_baseline}

\subsection{Tweets Dataset}
\label{sec:appendix_tw_baseline}

Table~\ref{tab:tw_baseline} shows all baseline F1-scores for Tweets dataset.

\begin{table*}[h]
    \centering
    \begin{tabular}{lccccccccc}
    \cline{2-10}
     & \multicolumn{9}{c}{\textbf{Augm. Type}} \\ \cline{2-10} 
    \textbf{} & \multicolumn{3}{c}{\textbf{EDA}} & \multicolumn{3}{c}{\textbf{Syn}} & \multicolumn{3}{c}{\textbf{BT}} \\ \cline{2-10} 
     & \multicolumn{9}{c}{\textbf{Augm.Perc.}} \\ \hline
    \textbf{Subset Size} & \textbf{0.05} & \textbf{0.1} & \textbf{0.2} & \textbf{0.05} & \textbf{0.1} & \textbf{0.2} & \textbf{0.05} & \textbf{0.1} & \textbf{0.2} \\ \hline
    
    500 & 0.83 & 0.82 & 0.83 & 0.83 & 0.83 & 0.83 & 0.87 & 0.87 & 0.87 \\
    1000 & 0.80 & 0.79 & 0.80 & 0.80 & 0.80 & 0.80 & 0.80 & 0.80 & 0.80 \\
    2000 & 0.81 & 0.81 & 0.81 & 0.79 & 0.79 & 0.79 & 0.78 & 0.78 & 0.78 \\
    5000 & 0.79 & 0.80 & 0.80 & 0.78 & 0.78 & 0.78 & 0.79 & 0.79 & 0.79 \\
    10000 & 0.78 & 0.78 & 0.77 & 0.77 & 0.78 & 0.77 & 0.78 & 0.78 & 0.78 \\
    \hline
    \end{tabular}
    \caption{Baseline F1-scores for Tweets dataset.}
    \label{tab:tw_baseline}
\end{table*}

\subsection{B2W Dataset}
\label{sec:appendix_b2w_baseline}

Table~\ref{tab:b2w_baseline} shows all baseline F1-scores for B2W dataset.

\begin{table*}[h]
    \centering
    \begin{tabular}{lccccccccc}
    \cline{2-10}
     & \multicolumn{9}{c}{\textbf{Augm. Type}} \\ \cline{2-10} 
    \textbf{} & \multicolumn{3}{c}{\textbf{EDA}} & \multicolumn{3}{c}{\textbf{Syn}} & \multicolumn{3}{c}{\textbf{BT}} \\ \cline{2-10} 
     & \multicolumn{9}{c}{\textbf{Augm.Perc.}} \\ \hline
    \textbf{Subset Size} & \textbf{0.05} & \textbf{0.1} & \textbf{0.2} & \textbf{0.05} & \textbf{0.1} & \textbf{0.2} & \textbf{0.05} & \textbf{0.1} & \textbf{0.2} \\ \hline
    
    500 & 0.92 & 0.92 & 0.91 & 0.94 & 0.94 & 0.94 & 0.92 & 0.92 & 0.92 \\
    1000 & 0.95 & 0.95 & 0.95 & 0.94 & 0.93 & 0.94 & 0.94 & 0.93 & 0.93 \\
    2000 & 0.93 & 0.94 & 0.93 & 0.93 & 0.93 & 0.93 & 0.94 & 0.94 & 0.93 \\
    5000 & 0.93 & 0.94 & 0.93 & 0.93 & 0.93 & 0.93 & 0.94 & 0.94 & 0.93 \\
    10000 & 0.94 & 0.94 & 0.93 & 0.94 & 0.94 & 0.94 & 0.93 & 0.93 & 0.93 \\
    \hline
    \end{tabular}
    \caption{Baseline F1-scores for B2W dataset.}
    \label{tab:b2w_baseline}
\end{table*}

\subsection{Mercado Libre Dataset}
\label{sec:appendix_ml_baseline}

Table~\ref{tab:ml_baseline} shows all baseline F1-scores for Mercado Libre dataset.

\begin{table*}[h]
    \centering
    \begin{tabular}{lccccccccc}
    \cline{2-10}
     & \multicolumn{9}{c}{\textbf{Augm. Type}} \\ \cline{2-10} 
    \textbf{} & \multicolumn{3}{c}{\textbf{EDA}} & \multicolumn{3}{c}{\textbf{Syn}} & \multicolumn{3}{c}{\textbf{BT}} \\ \cline{2-10} 
     & \multicolumn{9}{c}{\textbf{Augm.Perc.}} \\ \hline
    \textbf{Subset Size} & \textbf{0.05} & \textbf{0.1} & \textbf{0.2} & \textbf{0.05} & \textbf{0.1} & \textbf{0.2} & \textbf{0.05} & \textbf{0.1} & \textbf{0.2} \\ \hline
    
    500 & 0.62 & 0.61 & 0.63 & 0.64 & 0.62 & 0.62 & 0.65 & 0.66 & 0.66 \\
    1000 & 0.73 & 0.73 & 0.74 & 0.69 & 0.69 & 0.69 & 0.71 & 0.72 & 0.71 \\
    2000 & 0.80 & 0.80 & 0.80 & 0.79 & 0.78 & 0.79 & 0.77 & 0.78 & 0.77 \\
    5000 & 0.85 & 0.85 & 0.85 & 0.85 & 0.84 & 0.85 & 0.84 & 0.84 & 0.84 \\
    10000 & 0.88 & 0.88 & 0.88 & 0.88 & 0.88 & 0.88 & 0.89 & 0.89 & 0.89 \\
    \hline
    \end{tabular}
    \caption{Baseline F1-scores for Mercado Libre dataset.}
    \label{tab:ml_baseline}
\end{table*}

\section{Detailed F1-score Gains}
\label{sec:appendix_gains}

\subsection{Tweets Dataset}
\label{sec:appendix_tw_gains}

\begin{figure}[h]
    \centering
    \includegraphics[scale=0.5]{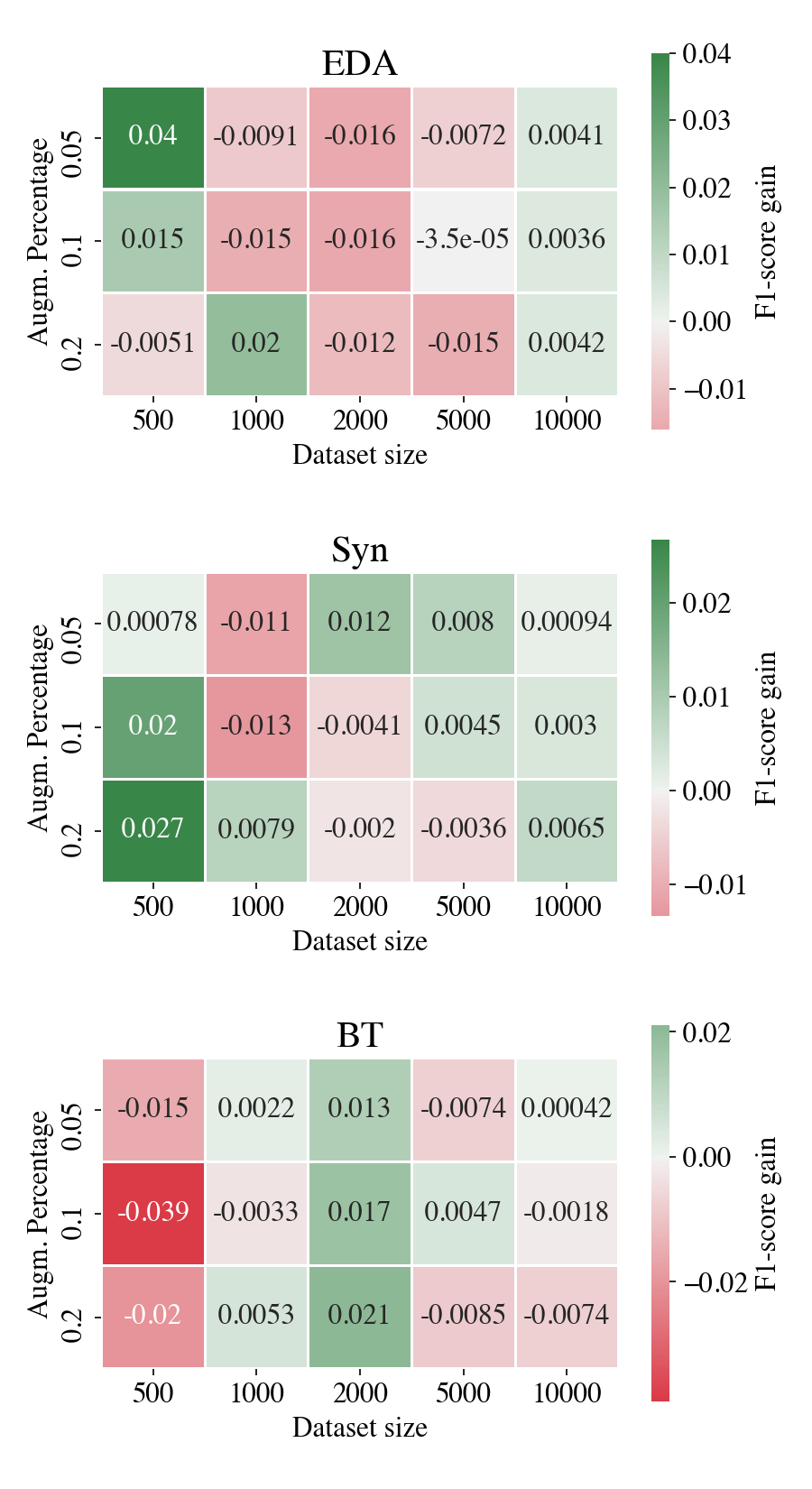}
    \caption{F1-score gains for each augmentation group on Tweets dataset.}
    \label{fig:tw_gain}
\end{figure}

Figure \ref{fig:tw_gain} shows all F1-score gains for Tweets dataset.

\subsection{B2W Dataset}
\label{sec:appendix_bw_gains}

\begin{figure}[h]
    \centering
    \includegraphics[scale=0.5]{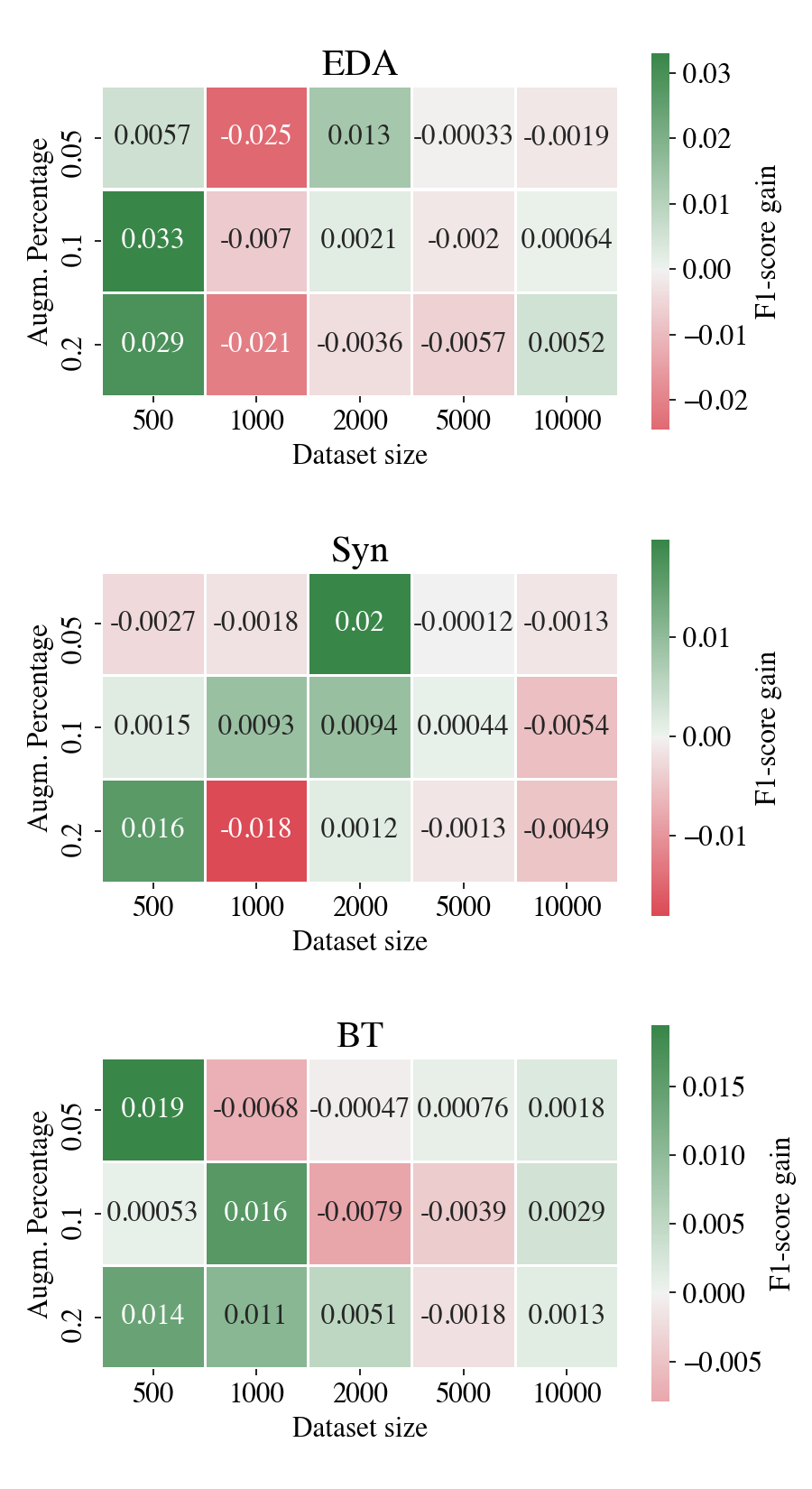}
    \caption{F1-score gains for each augmentation group on B2W dataset.}
    \label{fig:bw_gain}
\end{figure}

Figure \ref{fig:bw_gain} shows all F1-score gains for B2W dataset.

\subsection{Mercado Libre Dataset}
\label{sec:appendix_ml_gains}

\begin{figure}[h]
    \centering
    \includegraphics[scale=0.5]{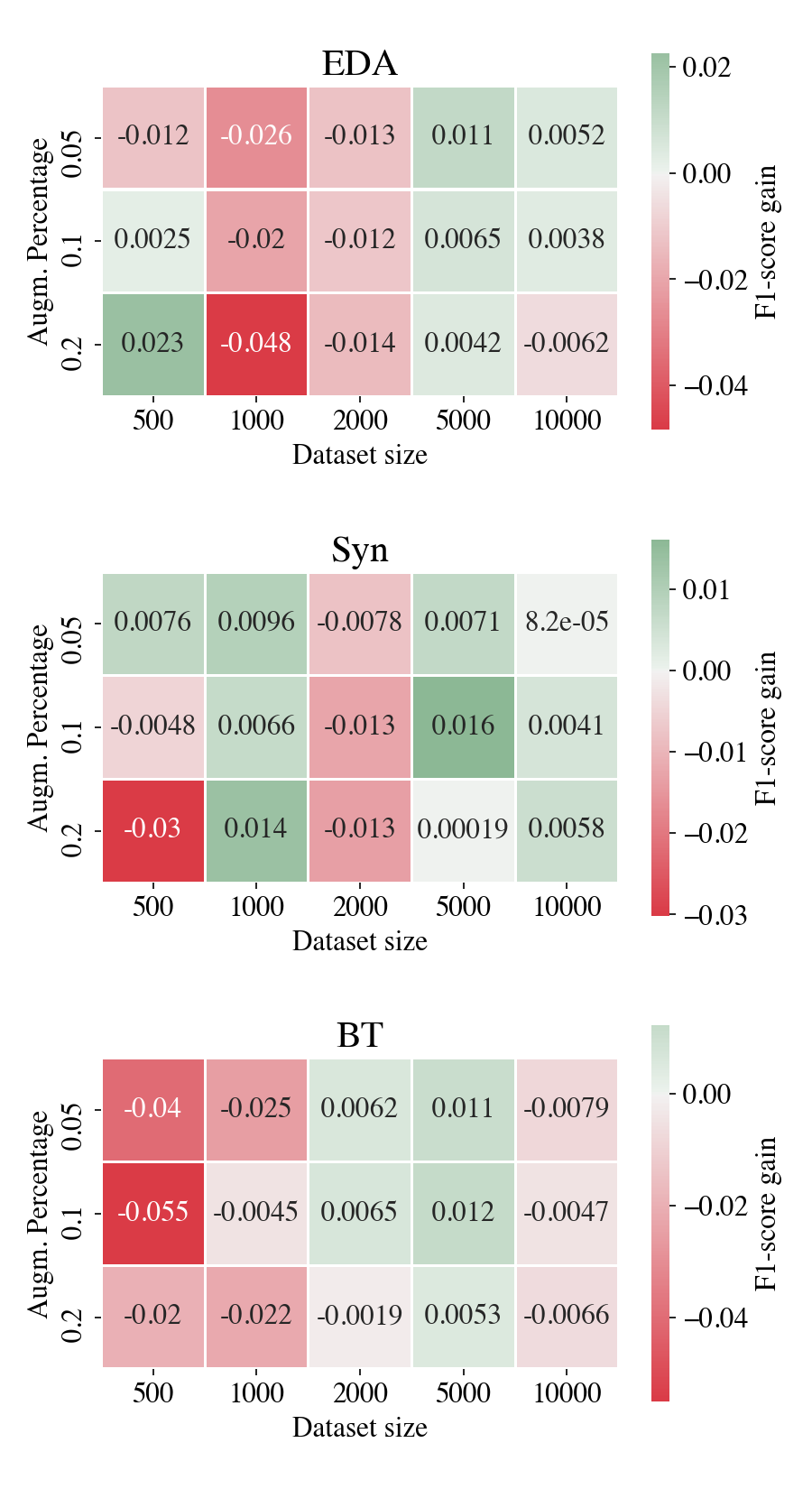}
    \caption{F1-score gains for each augmentation group on Mercado Libre dataset.}
    \label{fig:ml_gain}
\end{figure}

Figure \ref{fig:ml_gain} shows all F1-score gains for Mercado Libre dataset.

\section{Detailed \(p\)-values}
\label{sec:appendix_pvals}

\subsection{Tweets Dataset}
\label{sec:appendix_tw_pvals}

Table~\ref{tab:tw_pval} shows all \(p\)-values for Tweets dataset.

\begin{table*}[h]
\centering
\begin{tabular}{lccccccccc}
\cline{2-10}
 & \multicolumn{9}{c}{\textbf{Augm. Type}} \\ \cline{2-10} 
\textbf{} & \multicolumn{3}{c}{\textbf{EDA}} & \multicolumn{3}{c}{\textbf{Syn}} & \multicolumn{3}{c}{\textbf{BT}} \\ \cline{2-10} 
 & \multicolumn{9}{c}{\textbf{Augm. Perc.}} \\ \hline
\textbf{Subset Size} & \textbf{0.05} & \textbf{0.1} & \textbf{0.2} & \textbf{0.05} & \textbf{0.1} & \textbf{0.2} & \textbf{0.05} & \textbf{0.1} & \textbf{0.2} \\ \hline
500 & 0.30 & 0.89 & 1.00 & 1.00 & 0.67 & 0.60 & 0.70 & 0.24 & 0.58 \\
1000 & 0.81 & 0.61 & 0.62 & 0.74 & 0.68 & 0.87 & 0.93 & 1.00 & 0.86 \\
2000 & 0.43 & 0.43 & 0.63 & 0.60 & 0.95 & 0.95 & 0.63 & 0.48 & 0.32 \\
5000 & 0.53 & 0.97 & 0.24 & 0.54 & 0.78 & 0.67 & 0.55 & 0.81 & 0.50 \\
10000 & 0.61 & 0.58 & 0.52 & 0.91 & 0.81 & 0.35 & 0.98 & 0.88 & 0.45 \\
\hline
\end{tabular}
\caption{McNemar's test \(p\)-values for Tweets dataset.}
\label{tab:tw_pval}
\end{table*}

\subsection{B2W Dataset}
\label{sec:appendix_bw_pvals}

Table~\ref{tab:bw_pval} shows all \(p\)-values for B2W dataset.

\begin{table*}[h]
\centering
\begin{tabular}{lccccccccc}
\cline{2-10}
 & \multicolumn{9}{c}{\textbf{Augm. Type}} \\ \cline{2-10} 
\textbf{} & \multicolumn{3}{c}{\textbf{EDA}} & \multicolumn{3}{c}{\textbf{Syn}} & \multicolumn{3}{c}{\textbf{BT}} \\ \cline{2-10} 
 & \multicolumn{9}{c}{\textbf{Augm. Perc.}} \\ \hline
\textbf{Subset Size} & \textbf{0.05} & \textbf{0.1} & \textbf{0.2} & \textbf{0.05} & \textbf{0.1} & \textbf{0.2} & \textbf{0.05} & \textbf{0.1} & \textbf{0.2} \\ \hline
500 & 1.00 & 0.23 & 0.21 & 1.00 & 1.00 & 0.61 & 0.54 & 1.00 & 0.70 \\
1000 & 0.09 & 0.73 & 0.21 & 1.00 & 0.61 & 0.32 & 0.74 & 0.30 & 0.47 \\
2000 & 0.14 & 0.89 & 0.81 & \textbf{0.04} & 0.39 & 1.00 & 1.00 & 0.55 & 0.61 \\
5000 & 0.92 & 0.78 & 0.35 & 0.92 & 1.00 & 0.93 & 1.00 & 0.48 & 0.79 \\
10000 & 0.56 & 0.89 & 0.12 & 0.77 & 0.10 & 0.15 & 0.60 & 0.42 & 0.79\\
\hline
\end{tabular}
\caption{McNemar's test \(p\)-values for B2W dataset.}
\label{tab:bw_pval}
\end{table*}

\subsection{Mercado Libre Dataset}
\label{sec:appendix_ml_pvals}

Table~\ref{tab:ml_pval} shows all \(p\)-values for Mercado Libre dataset.

\begin{table*}[h]
\centering
\begin{tabular}{lccccccccc}
\cline{2-10}
 & \multicolumn{9}{c}{\textbf{Augm. Type}} \\ \cline{2-10} 
\textbf{} & \multicolumn{3}{c}{\textbf{EDA}} & \multicolumn{3}{c}{\textbf{Syn}} & \multicolumn{3}{c}{\textbf{BT}} \\ \cline{2-10} 
 & \multicolumn{9}{c}{\textbf{Augm. Perc.}} \\ \hline
\textbf{Subset Size} & \textbf{0.05} & \textbf{0.1} & \textbf{0.2} & \textbf{0.05} & \textbf{0.1} & \textbf{0.2} & \textbf{0.05} & \textbf{0.1} & \textbf{0.2} \\ \hline
500 & 0.93 & 0.93 & 0.80 & 1.00 & 0.93 & 0.50 & 0.60 & 0.44 & 0.81 \\
1000 & 0.79 & 1.00 & 0.36 & 1.00 & 0.95 & 0.66 & 0.69 & 0.80 & 0.95 \\
2000 & 0.52 & 0.55 & 0.54 & 0.75 & 0.67 & 0.56 & 0.67 & 0.67 & 0.96 \\
5000 & 0.63 & 0.62 & 1.00 & 0.72 & 0.13 & 0.93 & 0.56 & 0.44 & 0.81 \\
10000 & 0.51 & 0.66 & 0.23 & 0.89 & 0.72 & 0.60 & 0.23 & 0.38 & 0.21\\
\hline
\end{tabular}
\caption{McNemar's test \(p\)-values for Mercado Libre dataset.}
\label{tab:ml_pval}
\end{table*}

\end{document}